\documentclass[conference]{IEEEtran}
\IEEEoverridecommandlockouts
\usepackage{cite}
\usepackage{amsmath,amssymb,amsfonts}
\usepackage{algorithmic}
\usepackage{graphicx}
\usepackage{textcomp}
\usepackage{xcolor}
\usepackage{subcaption}
\usepackage{float}

\def\BibTeX{{\rm B\kern-.05em{\sc i\kern-.025em b}\kern-.08em
    T\kern-.1667em\lower.7ex\hbox{E}\kern-.125emX}}
\begin{document}

\title{Data Driven Approach to Input Shaping for Vibration Suppression in a Flexible Robot Arm}

\author{Jarkko Kotaniemi$^{1}$, Janne Saukkoriipi$^{1}$, Shuai Li$^{1}$, Markku Suomalainen$^{1}$

\thanks{Part of the work has been carried out within the framework of the EUROfusion Consortium, funded by the European Union via the Euratom Research and Training Programme (Grant Agreement No 101052200 — EUROfusion).}
		\thanks{$^{1}$J. Kotaniemi, J. Saukkoriipi, S. Li and M. Suomalainen are with the VTT Technical Research Centre of Finland Ltd, Oulu, Finland.
			{\tt\small jarkko.kotaniemi@vtt.fi}} 
        % <-this % stops a space
	}

\maketitle

\begin{abstract}
This paper presents a simple and effective method for setting parameters for an input shaper to suppress the residual vibrations in flexible robot arms using a data-driven approach. The parameters are adaptively tuned in the workspace of the robot by interpolating previously measured data of the robot's residual vibrations. Input shaping is a simple and robust technique to generate vibration-reduced shaped commands by a convolution of an impulse sequence with the desired input command. The generated impulses create waves in the material countering the natural vibrations of the system. The method is demonstrated with a flexible 3D-printed robot arm with multiple different materials, achieving a significant reduction in the residual vibrations.
\end{abstract}


\section{Introduction}
Undesired residual vibrations occur after performing motions and have multiple negative effects on robots: they decrease accuracy, lower lifespan, and compromise the structural integrity \cite{Khan2020Sliding}. In soft robots the vibration amplitude is even higher than in traditional robots and can even lead to system instability \cite{SicilianoKhatib2008}, meaning that in most applications, vibration damping or suppression is needed. By eliminating these vibrations on the control side, the structure of a robot system (soft or rigid) can be made more lightweight and flexible, without losing the accuracy of the system. 

The benefits of building a more lightweight robot improve as the size of the robot increases. Building a rigid robot at larger scales means having to add a lot of structural support, which makes the robot comparatively heavier than a smaller robot with similar rigidity. Adding a heavier load, especially on longer manipulators, also brings out the flexibility of even a very rigid robot; as the load gets heavier, the robot will naturally flex more. Robots made intentionally soft are especially susceptible to vibrations due to their intended lack of rigidity \cite{Thuruthel2018InducedVibrations}. Whereas soft robots are especially useful for applications that require a gentle or careful operation, such as many physical human-robot interaction use cases or others where human users must be near the robot, the control of softer robots also becomes more complex, and is an area of increasing interest \cite{kiang2015review}. 

Traditionally, the design philosophy for robot arms emphasizes rigidity to ensure the precision and accuracy required for robotic operations. However, this approach often results in increased weight and higher costs. Input shaping enables the creation of less rigid and lighter structures that still maintain the necessary accuracy and precision. Besides the advantage of reduced costs, in some applications where heavy robots are undesirable, such as collaborative robots (cobots) or robots aboard mobile platforms and drones, the flexible structure enhances safety for human operators within the workspace while still achieving the required task accuracy.

Input shaping works by manipulating the original motion command signal \cite{singer1990preshaping}, effectively splitting it into multiple sub-signals and introducing specific time delays between them, based on the half-period of the unwanted frequency. By doing so, these delayed components destructively interfere with the vibrations at the structure's natural frequency, reducing or eliminating them. The added delays incur a short motion time penalty with a reduction of more than 95\% of the amplitude in the vibrations \cite{singer1990preshaping} and an increase in the stability of the structure. Multiple input shapers with different delays can also be chained together to affect different frequencies at the same time \cite{kang2024multi}. 

It is important to note that the proposed input shaping method is purely feed-forward, even if external sensors are required to find the natural frequency of the robot manipulator. This choice simplifies the hardware needed for continuous operation of the robot, as external sensors provide an extra point of failure. In the remainder of this paper, we consider comparisons mainly to methods that are in nature feed-forward, and the sensors presented are only meant for the parameter tuning phase. Moreover, input shaping only requires the "model" of the system in the weakest sense, namely, the natural frequency of the system is sufficient; thus, heavy modeling work can be avoided. 

This paper demonstrates that a 3D-printed soft robot manipulator system can be augmented with a simple data-driven adaptive parameter-tuning input shaper to achieve significant vibration reduction. It is possible to avoid the inherent uncertainty in attempting to model the physical response of the robot and instead focus on the ground truth determined by sensors measuring the motion of the robot. This is especially useful in the use of 3D-printed robots where the behavior can be dependent on the settings used in the printer and the different properties of materials made by unrelated manufacturers. The parameters can be computed in a few key positions and then interpolated between them \cite{CVITANIC2020PoseOptimization}. By measuring the acceleration of the tip of the robot and performing a Fourier transform, a frequency spectrum is created, where the natural frequencies of the different modes of vibration can be seen. The half-period of these frequencies is one of the parameters of the input shaper. The other parameter is the damping factor, which is the factor of reduction in the vibration magnitude in the time interval of the half-period. In weakly damped systems, the damping factor is approximately equal to 1.

\section{Related work}

Input shaping is only one possibility for feed-forward vibration suppression; traditional (for example, Finite Impulse Response (FIR)) filters and careful controller design have been used for this task already for decades for mechanical systems \cite{smith1957posicast}, and also for robotics for a long while \cite{economou2000robust}. However, comparisons have been made between digital filtering and input shaping with an indication of input shaping suppressing vibrations faster than notch or low-pass filters \cite{singhose2010reducing}. Even though the study is old, and new studies are showing good results for other methods as well (for example, "feedback" without extra sensors by estimating the motions \cite{ito2016state} or model-based control \cite{huang2024research}), input shaping is a promising tool in feed-forward methods to research, not requiring a complicated model of the robot or constant feedback.

Whereas input shaping technologies share similar algorithms, there are different ways of finding the necessary parameters, typically divided into model-based and data-driven. For simpler 1 Degree of Freedom systems, such as anti-sway control of 3-D gantry crane systems \cite{ahmad2009investigations}, model-based is often a good approach due to the simplicity of the system. There are also successful examples of the model-based approach for more complex systems: an elastic-dynamic model-based design of an adaptive input shaper was proposed for the use in flexible robot manipulators in \cite{solatges2017adaptive}, and an inertia-based approach to creating a dynamics estimator for vibration control was done on a UR5e industrial robot arm using time-varying input shaping technology \cite{thomsen2021control}. 

However, such traditional input shaping methods developed for linear time-invariant systems are theoretically unsuitable for robots with nonlinear dynamics where the movement of the body affects weight distribution, making the inertia configuration dependent on the joint angles. Consequently, the resonance frequency of each axis, influenced by the equivalent inertia and flexibility of that specific axis, varies with these angles. Existing approaches have generalized input shaping to accommodate this state-dependent situation using various techniques. For instance, some methods select the most robust input shaper parameters by employing Iterative Learning Control (ILR) \cite{paperA}, while others determine appropriate input shaper parameters in real-time based on outputs from a decision tree \cite{paperB} or a neural network \cite{paperD}. Moreover, techniques such as notch filtering can also be utilized to suppress vibrations in multi-axis robots. However, it is important to note that notch filters implemented as Infinite Impulse Response (IIR) filters require infinite time to suppress residual vibrations fully. In contrast, input shaping, which is fundamentally an FIR filter, can effectively suppress residual vibrations in a finite time frame. In this work, an interpolation-based technique is developed to adapt input shapers for vibration suppression in multi-axis robots. This approach is more optimal than the iterative learning-based techniques \cite{paperA}, as the parameters change based on the pose of the robot, and requires significantly less computation than solutions utilizing decision trees or neural networks \cite{paperB,paperD}.

\section{Robot Design}

We developed a flexible-link robotic arm to be used mainly as a payload for a drone designed for sample collection. Due to its lightweight and flexible design, the arm provided an ideal test platform for conducting measurements and experiments using the input shaping method. To present a more difficult use case for testing, we introduced structural variability by changing one of the links to a different softer material. Since the arm is 3D-printed with various materials, traditional modeling of the structure is challenging. The uncertainty of its printed components' material properties makes parameter identification through modeling for the input shaper even more difficult. The challenges make it a great candidate for a data-driven approach.

\begin{figure}[h!]
\centering
\includegraphics[width=0.75\columnwidth]{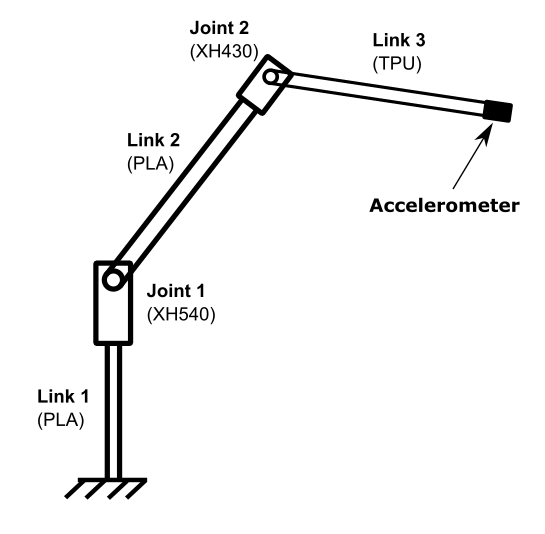}
\caption{\small{The robot arm structure on which the input shaping is tested with the joints and links specified.} }
\label{fig:structure}
\vspace{-0.7cm}
\end{figure}

The servos used were Dynamixel's XH430 and XH540 models, with the more powerful XH540 selected for the first joint and the faster, although weaker, XH430 chosen for the second joint. The first two joints were 3D printed from polylactic acid (PLA) to provide some flexibility while maintaining greater rigidity compared to the last link, which was 3D printed from the thermoplastic polyurethane (TPU). TPU, being more rubber-like, offers significantly higher flexibility than PLA and exhibits a completely different natural frequency. Additionally, the last link could accommodate extra weight to facilitate various test scenarios. The robot arm structure is shown in Figure  \ref{fig:structure}. 

An accelerometer was attached to the tip of the robot arm to measure the vibrations of the arm for parameter tuning, and it is not needed during runtime. 

\section{Input Shaping} 
\subsection{Control Scheme}
Input shaping is a technique for suppressing vibrations in lightly damped systems. We follow the work of  \cite{singer1990preshaping}. In the context of the flexible structure designed here, let the motor input signal be denoted by $u(t)$, and the system response by $y(t)$. The response $y(t)$ consists of two components: the desired robot movement $y_1(t)$ and the undesired vibration $y_2(t)$, so that
\begin{equation}
    y(t) = y_1(t) + y_2(t) \label{eq:response}
\end{equation}
If the input signal is delayed by a time $T_0$, the output will also be delayed by the same amount. Specifically, with input $u(t-T_0)$, the corresponding output is 
\begin{equation}
    y(t - T_0) = y_1(t - T_0) + y_2(t - T_0),\label{eq:delay}
\end{equation}
demonstrating the system’s time-invariant properties. Importantly, for vibrations, if we choose $T_0 = \frac{T}{2}$, where $T$ is the vibration period, the vibration component $y_2(t - T/2)$ will be in exact counter-phase with $ y_2(t)$, due to the symmetric nature of the oscillations.
Considering the damping effect, which reduces the amplitude of $y_2(t)$ over time, we can express this as 
\begin{equation}
    -k_0 \cdot y_2(t - T/2) = y_2(t),\label{eq:damping}
\end{equation}
where $k_0$ is the damping factor representing the reduction in vibration magnitude over a time interval of $T/2$. In weakly damped systems, this reduction is typically small, so $k_0$ is approximately equal to 1.
Thus, we can conclude that
\begin{eqnarray}
   && k_0 \cdot y(t - T/2) + y(t)\nonumber\\
     &=& k_0\cdot (y_1(t-T/2)+y_2(t-T/2)) + (y_1(t)+y_2(t))\nonumber\\
     &=& k_0 \cdot y_1(t - T/2) + y_1(t)+(k_0\cdot y_2(t-T/2)+y_2(t))\nonumber\\
     &=&k_0 \cdot y_1(t - T/2) + y_1(t).\label{eq:small_reduction}
\end{eqnarray}
Given that the system is linear near the operational point of the flexible structure, $ k_0 \cdot y(t - T/2) + y(t)$ corresponds to the response to the input signal 
\begin{equation}
    k_0 \cdot u(t - T/2) + u(t).\label{eq:input_signal}
\end{equation}
If the desired robot movement $y_1(t)$ is significantly slower than the vibration, we can reasonably assume that $ y_1(t - T/2) \approx y_1(t)$. This leads to the result that the input $ k_0 \cdot u(t - T/2) + u(t)$ produces a response of $(k_0 + 1) \cdot y_1(t)$. To achieve a pure, vibration-free response, we can apply the following shaped input $v(t)$:
\begin{equation}
    v(t) = \frac{k_0}{1 + k_0} \cdot u(t - T/2) + \frac{1}{1 + k_0} \cdot u(t).\label{eq:shaped_input}
\end{equation}

\begin{figure}[h!]
\centering
\includegraphics[width=\columnwidth]{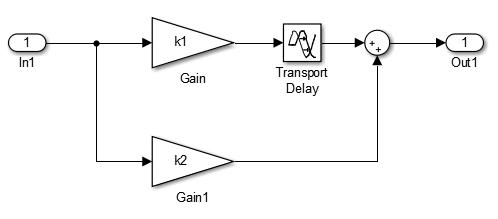}
\caption{\small{The implementation block diagram of the input shaper with $k_1=\frac{k_0}{1+k_0}$, $k_2=\frac{1}{1+k_0}$ and time delay for the half time of the vibration period.}}
\label{fig:block_diagram}
\vspace{-0.5cm}
\end{figure}

This input shaping can be easily implemented using the block diagram shown in Figure. \ref{fig:block_diagram}.

\begin{figure}[h!]
\centering
\includegraphics[width=\columnwidth]{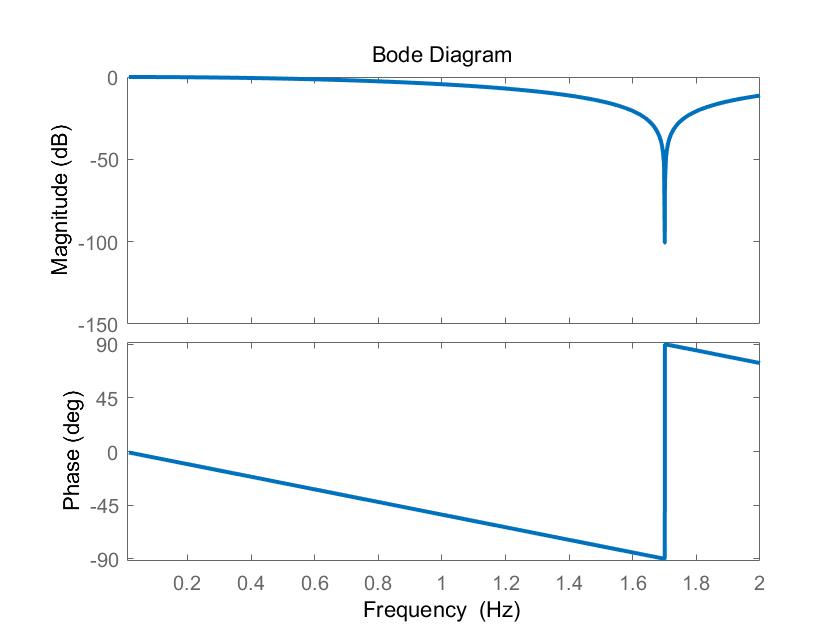}
\caption{\small{A Bode diagram shows there to be a clear suppression at $1.7$ Hz.} }
\label{fig:bode}
\vspace{-0.5cm}
\end{figure}

When the input shaper has been parameterized to reduce the frequency at $1.7$ Hz by setting the delay as the half period of the corresponding frequency, the frequency response of an input shaper is as shown in Figure \ref{fig:bode}. There is a clear suppression at that specific frequency.

\begin{figure}[h!]
\centering
\includegraphics[width=\columnwidth]{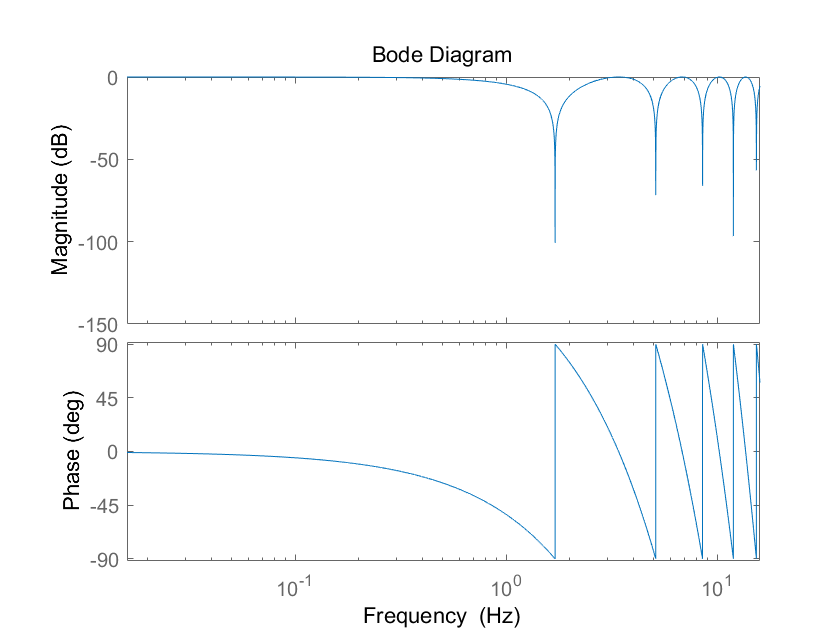}
\caption{\small{A Bode diagram of a wide range of frequencies showing the harmonics of the input shaper's frequency response.} }
\label{fig:bode_full}
\vspace{-0.5cm}
\end{figure}

The input shaper also reduces the frequencies at the other harmonics of the first frequency (Figure \ref{fig:bode_full}), but as the purpose of the input shaper in this use case is to only lower vibrations and not amplify any, the other harmonics do not cause any issue. 

\subsection{Parameter tuning}

For vibrational dampening, the parameters of the input shaper, $T_0$ and $k_0$, are chosen to match the natural vibrational frequencies and the damping factor of the system. A data-driven approach of measuring the frequency response of the system at a few key positions can be used to compute the parameters from the measurements. With the information gathered from a few positions within the workspace, the rest can be interpolated. The most important parameters to discover are the frequencies of the different modes of vibrations. The acceleration measurements of the vibrations after the actuated motion are fed through a discrete Fourier transform to get the frequency spectrum of the accelerations. The frequencies change gradually as the robot arm moves and its center of mass moves with it.

A proposed sequence for choosing the parameters of the input shaper is as follows:
\begin{enumerate}
    \item Choose key positions within the expected areas of the robot's operation.
    \item Gather acceleration data after moving to the chosen positions to measure the residual vibrations.
    \item Run a Fourier transformation to the measurements to get a frequency spectrum.
    \item Extract the peak frequencies from the spectrum as seen in Figure \ref{fig:2spikes}. The current method requires manual effort to extract the peaks, but an automated way is easily achievable.
    \item Correlate the frequencies to the positions to create a mapping as seen in Figure \ref{fig:fmap1}.
    \item Interpolate the frequencies to update the parameters of the input shapers when moving to a position within the map. The end position of the motion is used during the interpolation.
\end{enumerate}

The two main parameters of an input shaper are the time delay $T_0$ mentioned in Equation \ref{eq:delay}, and the damping factor $k_0$ mentioned in Equation \ref{eq:damping}. $T_0$ is also referred as $T/2$. By choosing these parameters correctly, in theory, the vibrations will be canceled entirely. In practice, finding the exact value is very difficult, as the system response is dependent on many outside factors, such as delays in the control system. Getting the values of the parameters close enough will still cause a great, but not total reduction, in vibrations. Using linear interpolation, the parameters can be computed at a reasonable accuracy even when the motion is going in a position between the previously chosen key positions. Extrapolating the parameters should also be possible as the mapping in Figures \ref{fig:fmap1} and \ref{fig:fmap2} show the change to seem at least locally linearizable. 

The chosen parameters are put into Equation \ref{eq:shaped_input}, which is also shown as a block diagram in Figure \ref{fig:block_diagram}. In a weakly damped system such as the one presented, $k_0$ can be assumed to be 1. $T/2$ is the half period of the natural frequency. The input signal of the motion is fed through the equation, and the output of that is sent to the servos. The resulting motion suppresses the residual vibrations. The input signal can be a velocity command or a position command. The actual speed of the motion does not matter. Any time a new motion is to be done, step 6 of the previous sequence is performed again.

\section{Implementation and Results}

\begin{figure}[h!]
\centering
\includegraphics[width=\columnwidth]{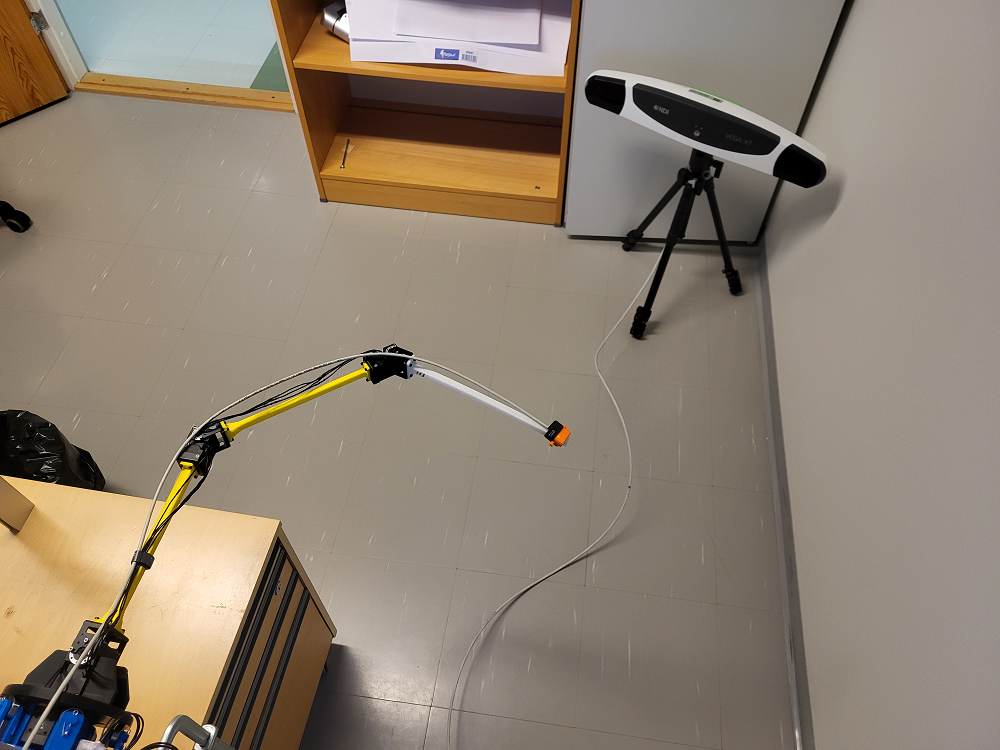}
\caption{\small{The experimental setup. The orange box at the tip of the robot arm is the IMU, and at the top right of the image is the laser tracker. Underneath the IMU is a laser reflector, which is followed by the tracker.} }
\label{fig:setup}
%\vspace{-0.5cm}
\end{figure}

Measurements for finding the natural frequency are done using a simple and cheap MPU6050 Inertial Measurement Unit (IMU) to measure the acceleration of the tip of the robot arm (Figure \ref{fig:setup}). The IMU does not need to be perfectly accurate as the measurements are transformed into the frequency domain, where the change in acceleration is important rather than the absolute value. To simplify measurements to ignore gravity, we positioned the robot arm and the IMU so that the measurements of the acceleration are done in the horizontal plane. This is to isolate the acceleration measurements to a specific axis of the IMU and so eliminate the contamination of unrelated vibrations in the measurements. An NDI Vega XT laser tracker was used to measure the position of the tip for verification of the technique. The measurements are done during and after the motion to show the residual vibrations.

The experimental setup is as follows: the robot arm is set to receive joint position commands for two different joints. First, the joint positions are sent to both joints without any modification as a step signal. The natural frequencies are calculated from the residual (motion after command ends) acceleration data in different positions and a map is created of the frequencies related to the positions where it was measured. Second, a position is chosen and the frequencies are interpolated from the frequency response map. This frequency is used as a parameter for the input shaper and the joint position step signal is fed through the input shaper. The vibrations are observed in this scenario and compared to a situation in the same position without the input shaper.

\begin{figure}[h!]
\centering
\includegraphics[width=\columnwidth]{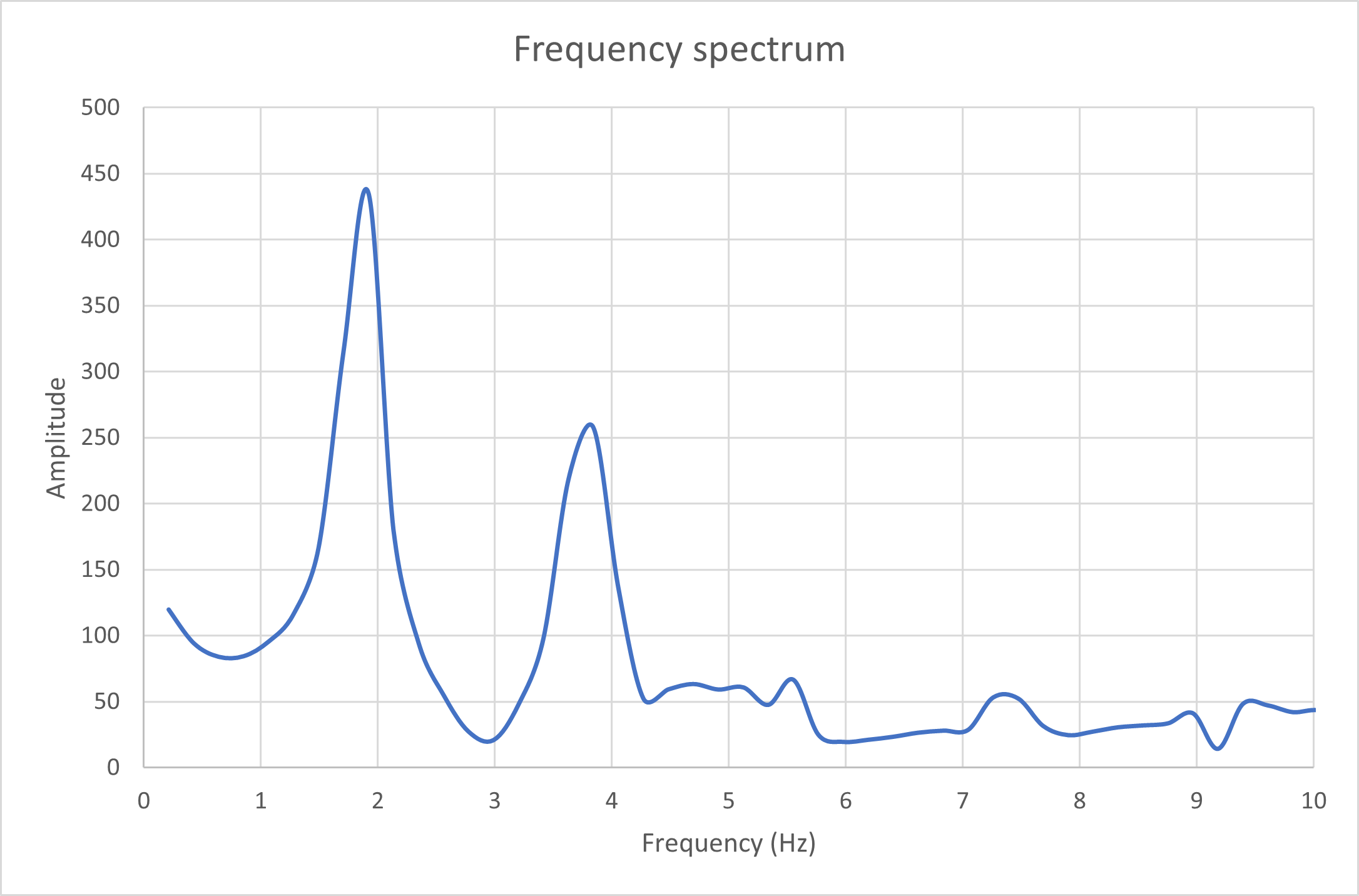}
\caption{\small{Two modes of vibration from the frequency spectrum. The first mode is 1.9 Hz and the second is 3.8 Hz. The position of the first joint is at 45$^{\circ}$ and the second at 60$^{\circ}$.} }
\label{fig:2spikes}
%\vspace{-0.5cm}
\end{figure}

Based on the experiments, there are two clearly distinct modes of vibration of the structure (Figure \ref{fig:2spikes}). The lowest frequency mode of vibration is the robot arm vibrating from the base and the second mode is the last link, which is made from a different material, vibrating from the last joint.

\begin{figure}[h!]%
  \centering
    \begin{subfigure}{0.33\columnwidth}
        \includegraphics[width=\columnwidth]{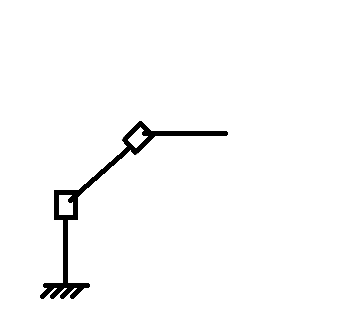}
         \caption{}
         \label{fig:poseA}
    \end{subfigure}\hfill%
    \begin{subfigure}{0.33\columnwidth}
        \includegraphics[width=\columnwidth]{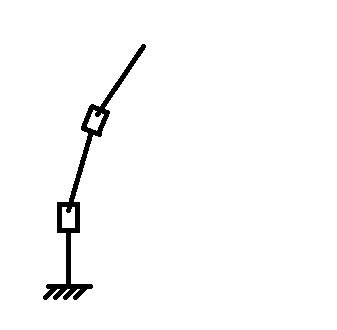}
        \caption{}\label{fig:poseB}
    \end{subfigure}\hfill%
    \begin{subfigure}{0.33\columnwidth}
        \includegraphics[width=\columnwidth]{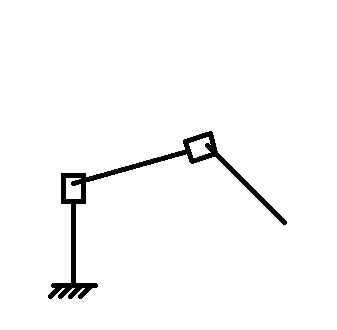}
         \caption{}\label{fig:poseC}
    \end{subfigure}\hfill%

\caption{Reference images of the named poses of the robot arm; a) position A joint values are 45$^{\circ}$ and 45$^{\circ}$; b) position B joint values are 15$^{\circ}$ and 15$^{\circ}$; c) position C joint values are 75$^{\circ}$ and 60$^{\circ}$.}
\label{fig:poses}
%\vspace{0.5cm}
\end{figure}

\begin{figure}[h!]
\centering
\includegraphics[width=\columnwidth]{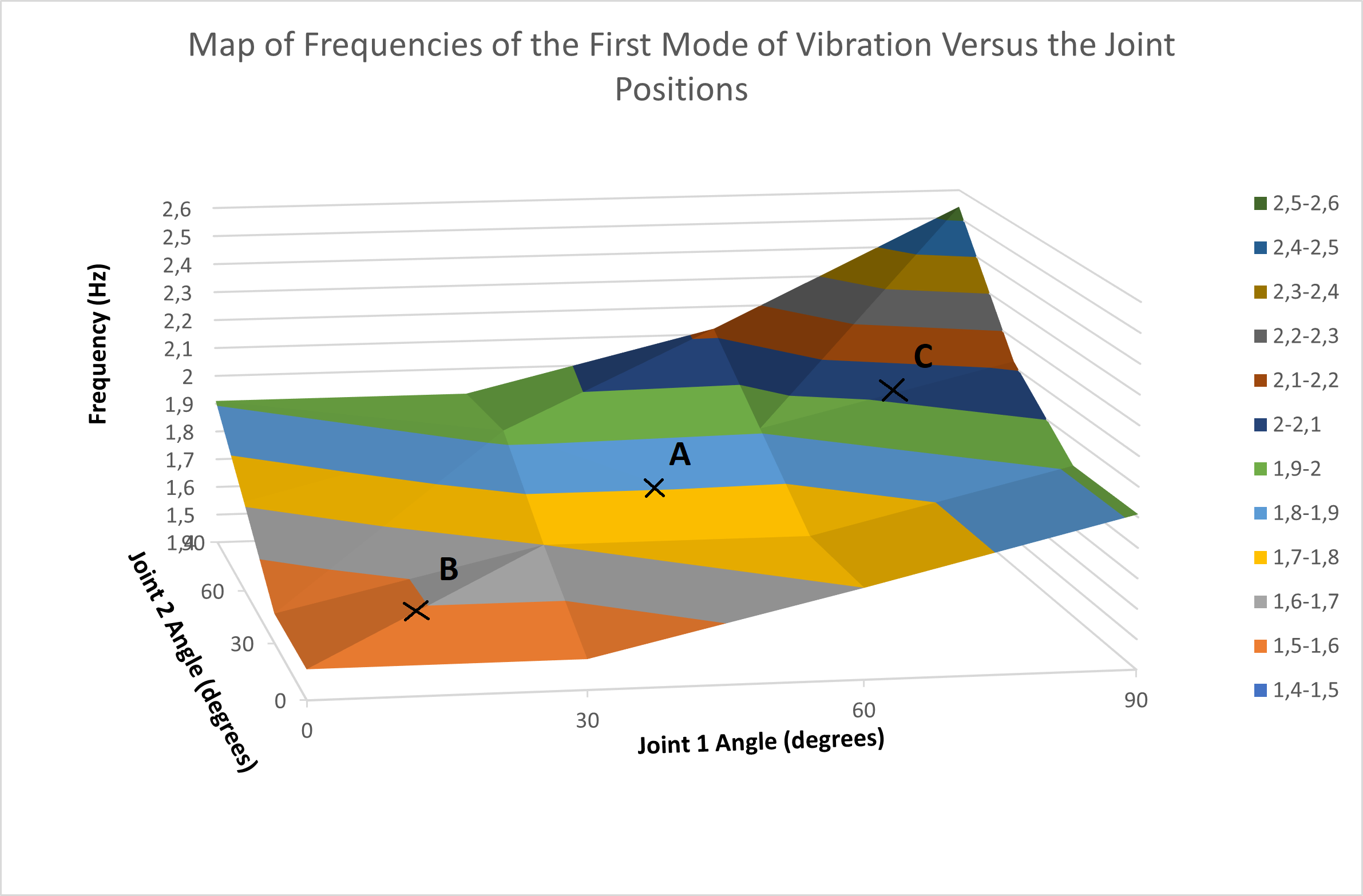}
\caption{\small{The frequency of the first mode of vibrations versus the joint positions. Positions A, B, and C are marked on the map with pose references.} }
\label{fig:fmap1}
%\vspace{-0.5cm}
\end{figure}

\begin{figure}[h!]
\centering
\includegraphics[width=\columnwidth]{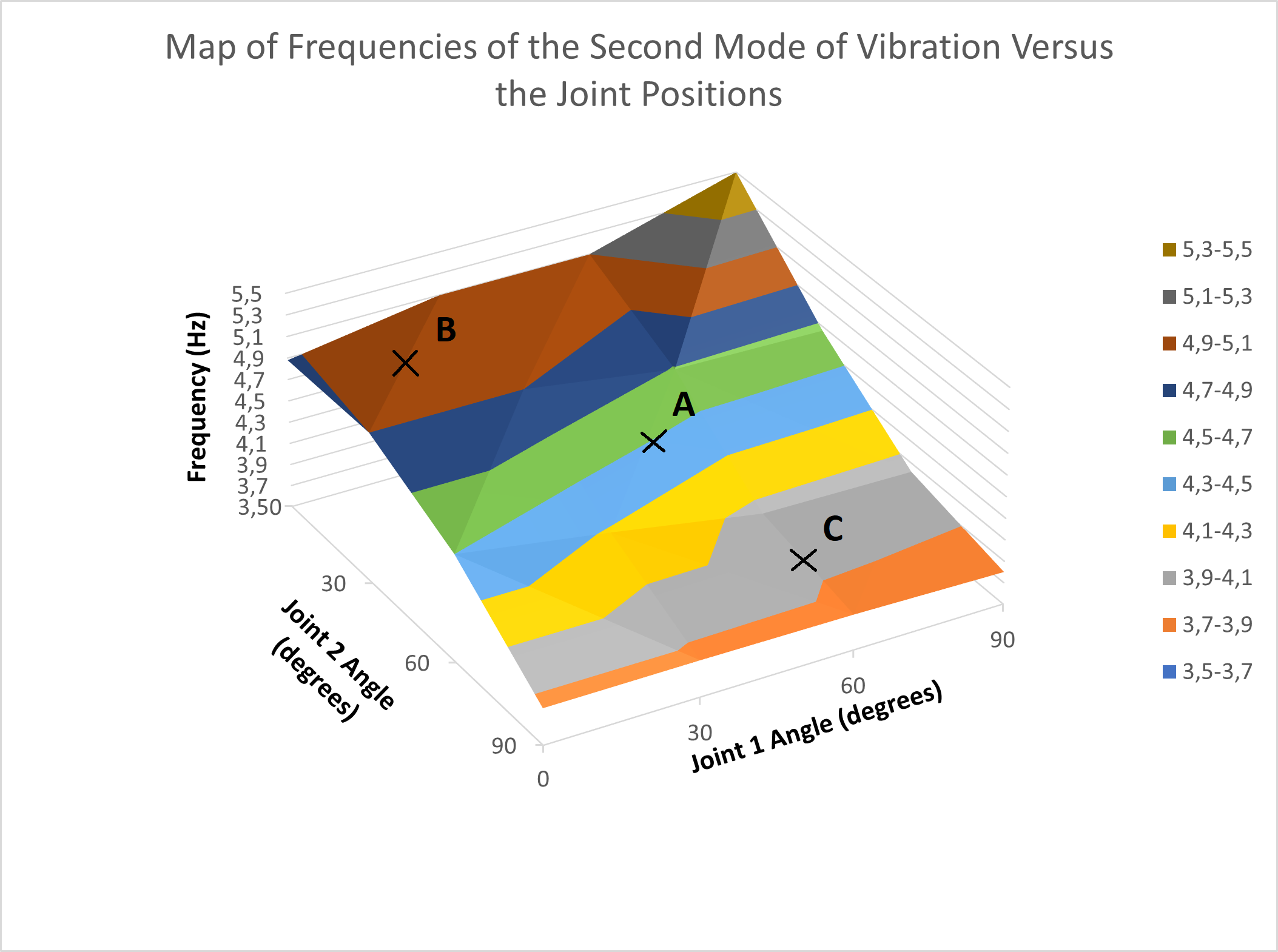}
\caption{\small{The frequency of the second mode of vibrations versus the joint positions. Joint 2 values are reversed compared to Figure \ref{fig:fmap1} for visualization purposes. Positions A, B, and C are marked on the map with pose references.} }
\label{fig:fmap2}
\vspace{-0.5cm}
\end{figure}

Measuring the frequencies of the two modes of vibration over a range of both joints' positions, from 0 degrees to 90 degrees with two other points in between for a total of 16 points, results in a map of frequencies for the two modes of vibration (figures \ref{fig:fmap1} and \ref{fig:fmap2}). As the robot arm is in different positions, the center of mass and the shape of the arm changes, which results in different frequencies of the modes of vibration. The frequency map can then be used to interpolate the frequencies in between the measured positions. In the maps, there are three positions marked. These refer to the end position of a motion where the frequency is measured. The joint values at position A are 45$^{\circ}$ and 45$^{\circ}$, at position B they are 15$^{\circ}$ and 15$^{\circ}$, finally at position C they are 75$^{\circ}$ and 60$^{\circ}$. The positions are visualized in Figure \ref{fig:poses}.

As there are two distinct modes of vibration, it follows that reducing both is necessary. By chaining two input shapers together, where the output of the first is fed as input into the second, both frequency peaks can be targeted. The trade-off in using multiple input shapers in this way is that the delay of the signal keeps increasing. 

\begin{figure}[t]
\centering
\includegraphics[width=\columnwidth]{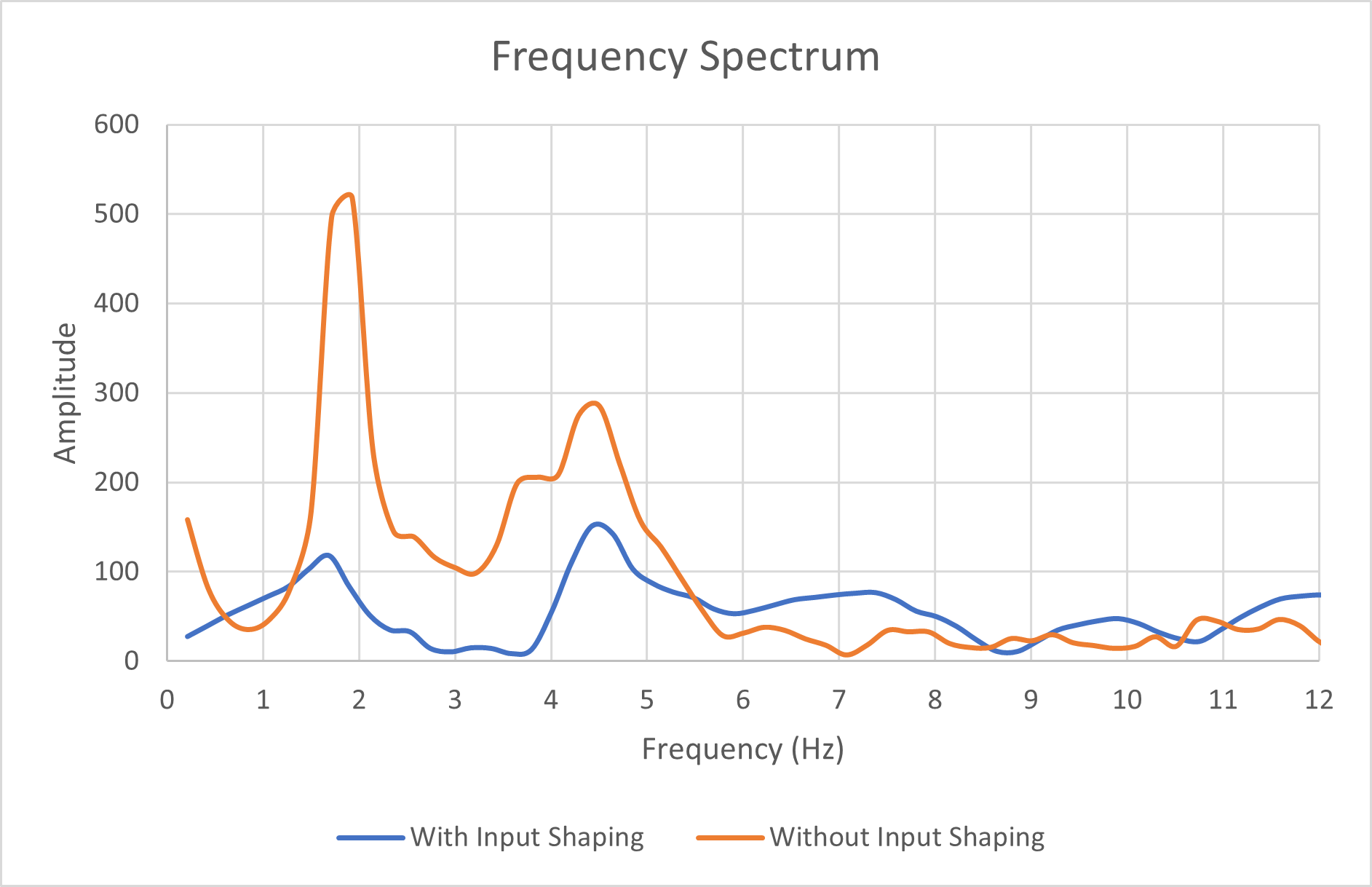}
\caption{\small{The change in frequency response with and without the input shaper. The robot arm performs a motion, where both joints go from 0$^{\circ}$ to 45$^{\circ}$ (position A).} }
\label{fig:freq_change}
\vspace{-0.5cm}
\end{figure}

The input shaper reduces the amplitude of the vibrations in the chosen frequencies and the corresponding reduction of the peaks is shown in the frequency spectrum as seen in Figure \ref{fig:freq_change}. There are still some residual vibrations left mostly in the higher frequency mode, but the reduction is still significant. Experiments show that the reduction can be up to 97\% as seen in Table \ref{tab:table1}.

\begin{figure}[h!]
\centering
\includegraphics[width=\columnwidth]{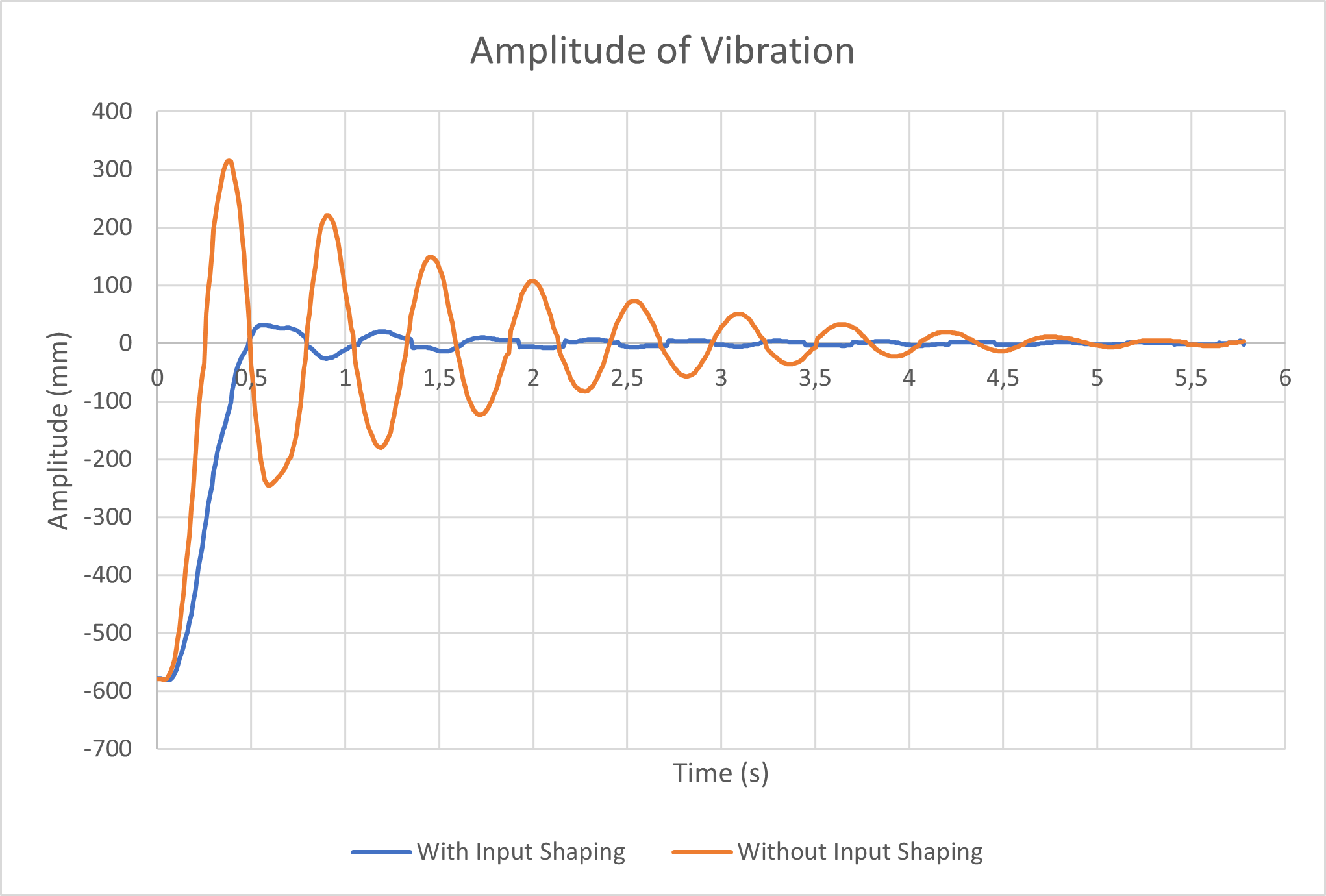}
\caption{\small{Amplitude measurement with the laser tracker of the robot arm's tip over time, with and without the input shaper. The robot arm performs a motion, where both joints go from 0$^{\circ}$ to 45$^{\circ}$ (position A).} }
\label{fig:laser}
\vspace{-0.5cm}
\end{figure}

The amplitude of the vibrations measured by the laser tracker at position A is seen in Figure \ref{fig:laser} with a comparison of the measurements using an input shaper and not using one.

\begin{table}[h!]
    \centering
    \caption{The reduction in the amplitude of vibration.}
    \begin{tabular}{|c|c|c|c|}
        \hline
        Position & Amplitude without IS & Amplitude with IS & Reduction \\ 
        \hline
        A & 304 mm & 29.8 mm & 90.2\% \\ 
        \hline
        B & 387 mm & 11.1 mm & 97.1\% \\ 
        \hline
        C & 218 mm & 36.4 mm & 83.3\% \\
        \hline
    \end{tabular}
    \label{tab:table1}
    \vspace{-0.5cm}
\end{table}

\section{Conclusion}
This paper presents a data-driven solution for input shaper parametrization for the purposes of vibration suppression with adaptive tuning. It is a robust and effective solution for the usage in unconventional flexible robot arms. It is demonstrated on a flexible 3D-printed robot arm with multiple different materials. 

For future work, the same procedure is extended to further joints and more dimensions. A more complete method of analyzing the acceleration data in three dimensions is necessary for future work of adding more degrees of freedom and moving and rotating the tip of the robot in all the spatial dimensions. As the modes of vibration are also going to be in different directions, it is important to correlate the effects of different joints' movements to the vibrations caused by that movement. Measuring all the possible ways a robot arm with six degrees of freedom can vibrate and coming up with a way to cancel all of them will be in the future. Methods of updating the performance of the input shapers during operation and attempting to extrapolate the parameters outside the measured areas will be considered. Based on the related work, some iterative learning methods already exist, which could be integrated into this system.

This current work is used in small-size robot arms, but the same method should work even in very large systems. The research will focus on controlling manipulators carrying multi-ton objects, where the inherent size and weight make even robust systems flexible.

\vspace{-0.1cm}

\bibliographystyle{unsrt}
\bibliography{biblio.bib}

\end{document}